\documentclass[]{spie}

\usepackage{geometry}
\usepackage{amssymb}
\usepackage{pifont}

\usepackage{booktabs}
\usepackage{siunitx}
\usepackage{amsmath}
\usepackage{adjustbox}
\usepackage{amsmath,amsfonts,amssymb}
\usepackage{graphicx}
\usepackage{cite} 
\usepackage{times}
\usepackage{epsfig}
\usepackage{amsmath}
\usepackage{nccmath}
\usepackage{amssymb}
\usepackage{mwe}
\usepackage{acro}
\usepackage{amssymb}
\usepackage{xcolor,colortbl}
\usepackage{tabularx}
\usepackage{relsize}
\usepackage{pifont}
\usepackage{booktabs} 
\usepackage{multirow}
\usepackage{multicol}
\usepackage{adjustbox}
\usepackage{makecell}
\usepackage{multirow}
\usepackage{float}
\usepackage{graphicx}
\usepackage{makecell}
\usepackage{tabu}
\usepackage[colorlinks=true, allcolors=blue]{hyperref}
\usepackage[capitalize]{cleveref}

\title{Seeing Abnormal from Normal: Glomerular Abnormality in Representations of Normal Renal Morphology}

\author[a]{Greta Hasko}
\author[b]{Rachit Saluja}
\author[c]{Tianyu Shi}
\author[d]{Leiyue Zhao}
\author[e]{Yuechen Yang}
\author[f]{Daniel Reisenbuechler}
\author[e]{Tianyuan Yao}
\author[g]{Zhenhao Guo}
\author[h]{John Cannon}
\author[i]{Yuling Chi}
\author[i]{Lorraine Gudas}
\author[b,i]{Mert R. Sabuncu}
\author[i]{Yihe Yang}
\author[i]{Ruining Deng}

\affil[a]{Cornell University, Ithaca, NY 14853, USA}
\affil[b]{Cornell Tech, New York, NY 10044, USA}
\affil[c]{Sichuan University, Chengdu, 610207, CN}
\affil[d]{Johns Hopkins University, Baltimore, MD 21218, USA}
\affil[e]{Vanderbilt University, Nashville, TN, 37235, USA}
\affil[f]{University of Regensburg, Regensburg, Bavaria 93053, DE}
\affil[g]{New York University, New York, NY 10012, USA}
\affil[h]{New York Medical College, Valhalla, NY, 10595, USA}
\affil[i]{Weill Cornell Medicine, New York, NY 10065, USA}

\begin{document} 
\maketitle
\enlargethispage{2\baselineskip}
\begingroup
\setlength{\parindent}{0pt}
\setlength{\hangindent}{0pt}
\setlength{\leftskip}{0pt}
\setlength{\rightskip}{0pt}
\renewcommand{\footnoterule}{%
  \noindent\rule{\textwidth}{0.4pt}\par
}
\footnotetext{%
  \makebox[\textwidth][l]{Further author information: (Send correspondence to Ruining Deng)}\\
  \makebox[\textwidth][l]{\hspace*{1.5em}E-mail: rud4004@med.cornell.edu}%
}
\endgroup

\begin{abstract}
Fine-grained evaluation of glomerular pathology must distinguish normal glomeruli from abnormalities such as global and segmental glomerulosclerosis, obsolescent, ischemic, solidified, disappearing, and atubular glomeruli. Supervised classification requires labeled examples of every category, which is impractical when subtypes are rare, heterogeneous, or absent from the training cohort. One-class anomaly detection offers an alternative by modeling normal data and scoring deviations, allowing previously unseen abnormalities to be detected. We use the frozen residual U-Net backbone of Omni-Seg, pretrained to segment structurally normal renal primitives without abnormal-subtype labels. We propose \emph{NoRDeC} (Normal-Reference Detection and Characterization), a framework combining Mahalanobis normal-reference scoring with layer-wise representation analysis to determine whether and where glomerular pathology is encoded, how spatial aggregation affects detection, and whether abnormalities alter inter-layer relationships differently. Using glomerular images from two institutions, we evaluate backbone layers and aggregation strategies, compare NoRDeC with PaDiM and PatchCore, and analyze representations using centered kernel alignment (CKA). Layer 4 with Center-70 aggregation achieved a pooled AUROC of $0.926\pm0.013$. NoRDeC achieved the highest AUROC in six of seven abnormality categories and in the pooled analysis, while CKA suggested subtype-dependent changes in inter-layer relationships not captured by anomaly scores alone. The normal-reference model is fitted using only normal glomeruli; abnormality labels are used for configuration selection, evaluation, and grouping in the representation analysis. These results demonstrate that a frozen renal feature extractor can support both detection and representation-level characterization of glomerular abnormalities without using abnormal examples to fit the detector.
\end{abstract}


\section{Description of purpose}
\label{sec:intro} 
\begin{figure}
    \centering
    \includegraphics[width=0.8\linewidth]{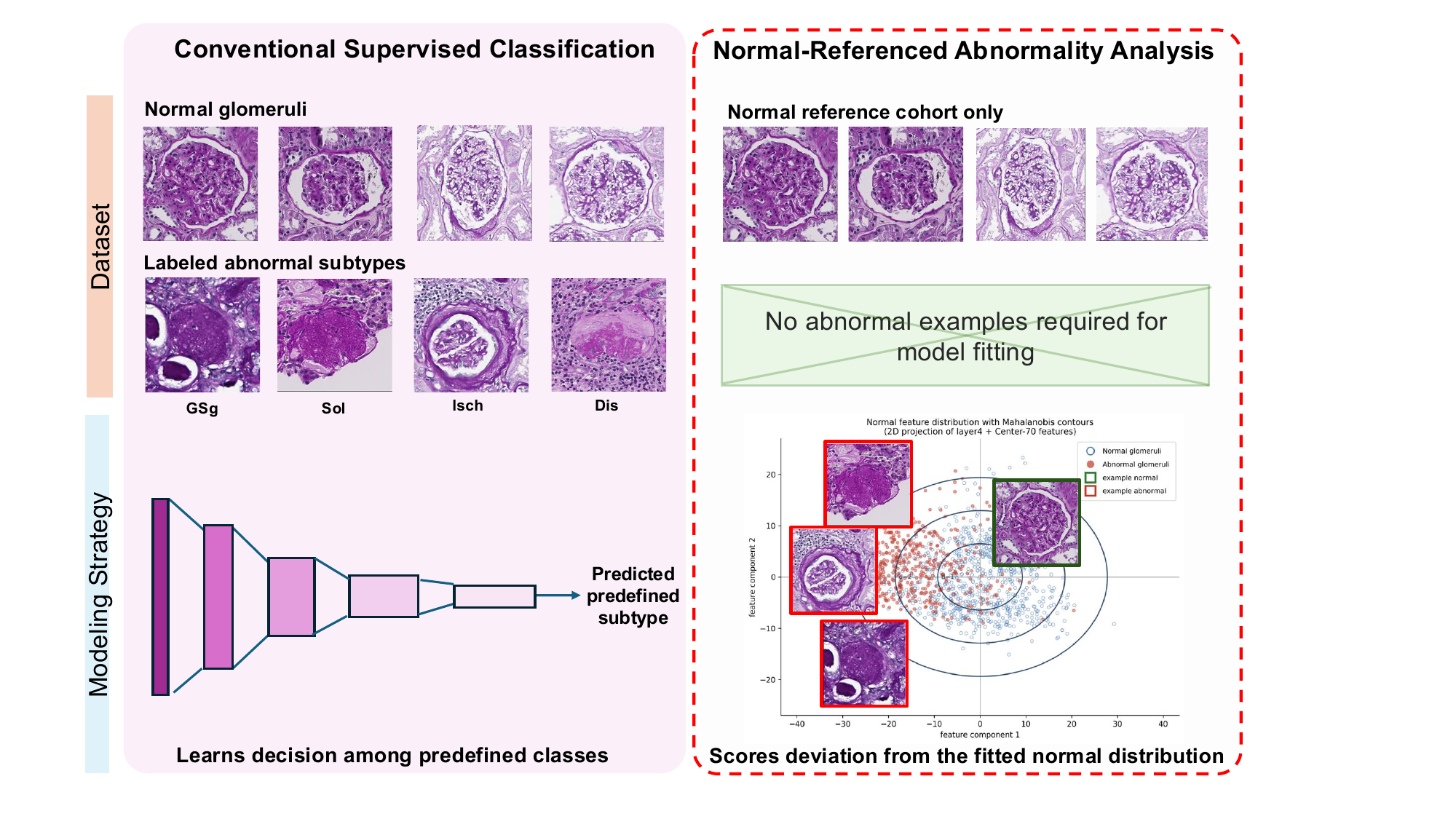}
    \caption{\textbf{Conceptual comparison between conventional supervised classification and the proposed normal-reference abnormality analysis.} Conventional supervised classification requires labeled normal glomeruli and predefined abnormal subtypes and predicts among the classes represented during training. In contrast, the proposed approach fits a normal feature distribution using only the normal reference cohort and assigns each test glomerulus an abnormality score according to its deviation from that distribution. Abnormal glomeruli are used only for evaluation and representation analysis, not for fitting the normal-reference model. The two-dimensional feature projection is provided for visualization.}
    \label{fig:problem}
\end{figure}

Fine-grained characterization of glomerular morphology is an important component of computational kidney pathology~\cite{he2026glopath,bouteldja2020deep,nan2022automatic,deng2024prpseg,deng2024hats}. Beyond the binary distinction between normal and abnormal glomeruli, this study considers seven abnormal categories: global glomerulosclerosis (GSg), segmental glomerulosclerosis (SGS), obsolescent, solidified (Sol), ischemic (Isch), disappearing (Dis), and atubular glomeruli. These categories differ substantially in their morphology, prevalence, and pathological significance~\cite{mangrum2024disruption,ancajas2023cellular,denic2019glomerular}. Compared with normal glomeruli, which exhibit relatively homogeneous morphology, abnormal glomeruli are often rare, heterogeneous, and unevenly represented across diseases and institutions~\cite{ling2026protective,schaub2023spatial,puelles2011glomerular}. As illustrated in Fig.~\ref{fig:problem}, supervised fine-grained classifiers require sufficient labeled examples of every target category~\cite{yao2022self,yu2025glo,guo2025classification}. Assembling such datasets is impractical for uncommon abnormalities, and closed-set classifiers cannot explicitly recognize categories that are absent from the training data. These challenges have been demonstrated in glomerular pathology. In a recent study of kidney biopsies collected from 22 institutions, crescents and segmental sclerosis accounted for only 2\% to 3\% of the annotations. During cross-scanner evaluation, the crescent-detection $\mathrm{AP}_{50}$ decreased from 0.64 to 0.19, corresponding to a relative reduction of 70.3\%~\cite{matsui2025domain}.

One-class anomaly detection addresses this challenge by constructing a reference model from normal samples and identifying observations that deviate from the learned normal representation, thereby avoiding the need for abnormality-specific supervision during model fitting~\cite{cui2023feasibility,cui2025quantitative,huo2021ai}. This approach is particularly suitable for computational pathology, where normal morphology is generally better represented in available datasets, whereas pathological variation is diverse and often sparsely sampled. Feature-based anomaly-detection methods, including Patch Distribution Modeling (PaDiM)~\cite{defard2021padim} and PatchCore~\cite{roth2022towards}, compare test-image features with statistical models or representative feature collections constructed from normal training images. PaDiM models the distribution of normal features, whereas PatchCore compares test features with a memory bank of representative normal features. Although these methods have demonstrated strong anomaly-detection and localization performance, they are designed primarily to produce anomaly scores and localization maps. They do not explicitly characterize how abnormality-related information changes across the network hierarchy. Therefore, it remains unclear how pathological structures are encoded within representations learned without abnormality supervision, at which layers abnormality-related information is most strongly expressed, and whether different pathological processes produce distinct changes in these representations.

In this work, we introduce Normal-Reference Detection and Characterization (NoRDeC), a framework that combines normal-reference anomaly detection with representation analysis to determine not only whether glomerular abnormalities can be detected but also how they are encoded in features learned without abnormality-specific supervision. NoRDeC uses (i) multi-layer embeddings extracted from the frozen Omni-Seg backbone~\cite{deng2023omni} with center-restricted spatial pooling, (ii) a Gaussian normal-reference model whose covariance is estimated using Ledoit and Wolf shrinkage~\cite{ledoit2004well} and whose abnormality scores are calculated using the Mahalanobis distance~\cite{mefleh2025beyond}, and (iii) pairwise centered kernel alignment (CKA) across backbone layers~\cite{kornblith2019similarity} to characterize abnormality-specific changes in inter-layer similarity patterns. Using glomerular datasets from two institutions, we demonstrate that intermediate-to-deep encoder features provide the strongest discrimination, center-restricted spatial pooling improves detection performance, and different glomerular abnormalities produce distinct inter-layer similarity patterns. Overall, these findings show that the frozen feature hierarchy contains meaningful information about glomerular pathology and can support both one-class abnormality detection and representation characterization.

\section{Methods}

\begin{figure}
    \centering
    \includegraphics[width=1\linewidth]{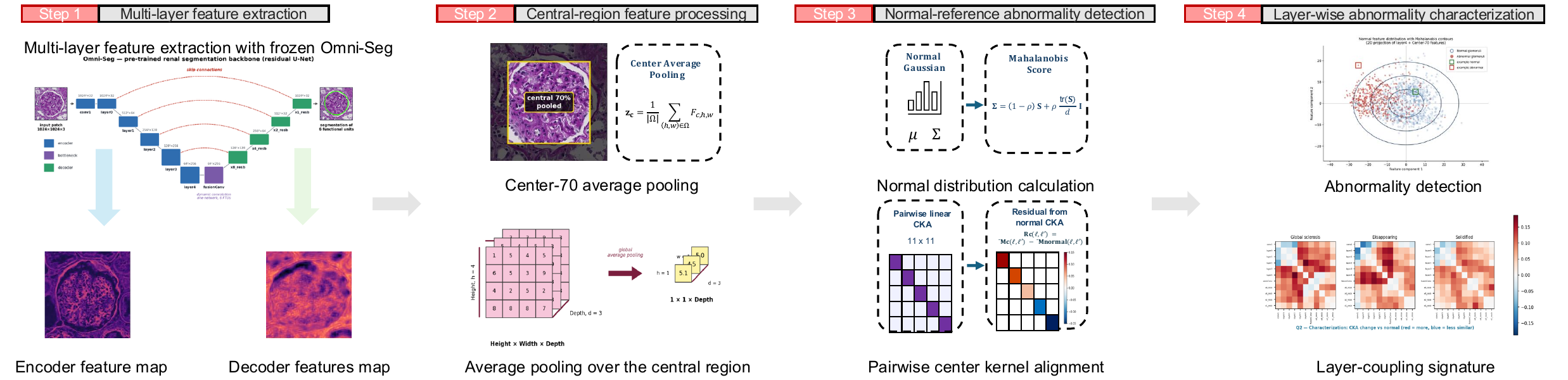}
    \caption{\textbf{Overview of the four components of NoRDeC.} Multi-layer feature maps are extracted from the frozen Omni-Seg backbone and processed within the central region. Center-70 pooled embeddings are used for Gaussian normal-reference modeling and Mahalanobis abnormality scoring. The corresponding unpooled spatial feature maps are used for pairwise CKA, and their deviations from the mean normal CKA matrix provide abnormality-specific layer-coupling signatures.}
    \label{fig:method_overview}
\end{figure}

The complete NoRDeC workflow is summarized in Fig.~\ref{fig:method_overview}. The workflow consists of four components. First, multi-layer feature maps are extracted from the frozen Omni-Seg backbone. Central-region processing then produces a pooled embedding for abnormality scoring while preserving the spatial feature maps required for representation analysis. The pooled embeddings are used to fit a Gaussian normal-reference model and calculate Mahalanobis abnormality scores. In parallel, the spatial feature maps are compared across network stages using CKA to derive abnormality-specific residual signatures relative to normal glomeruli. These two complementary analyses provide abnormality detection and representation characterization, respectively.

\subsection{Multi-layer feature extraction using frozen Omni-Seg}

Omni-Seg~\cite{deng2023omni} is a pretrained scale-aware dynamic segmentation network with a residual U-Net backbone for segmenting renal functional units at multiple magnifications. It was trained using renal-biopsy regions of interest from patients with minimal change disease without abnormality supervision, and its training subjects and slides do not overlap with those used in this study. We freeze the backbone and extract feature maps from eleven stages: $\mathrm{conv1}$, $\mathrm{layer0}$ through $\mathrm{layer4}$, $\mathrm{fusionConv}$, and $\mathrm{x8\_resb}$ through $\mathrm{x1\_resb}$, with channel dimensions of 32, 32, 64, 128, 256, 256, 256, 128, 64, 32, and 32, respectively. Because the task and scale codes are applied only in the downstream dynamic segmentation head, the extracted backbone features are independent of the requested segmentation task. For each $1024\times1024$ glomerular patch, stage $\ell$ produces a spatial feature map $F^{(\ell)}\in\mathbb{R}^{C_\ell\times H_\ell\times W_\ell}$.

\subsection{Central-region feature processing}

Each glomerulus is centered in its image patch, whereas the corners may contain surrounding non-glomerular tissue. We therefore define \emph{Center-70} as the central window containing $70\%$ of the feature-map height and $70\%$ of its width, corresponding to $49\%$ of the total area. For abnormality scoring, the feature map is average-pooled within this window according to Eq.~\eqref{eq:center70}:
\begin{equation}
\begin{aligned}
z_c^{(\ell)}
&=
\frac{1}{|\Omega_\ell|}
\sum_{(h,w)\in\Omega_\ell}
F_{c,h,w}^{(\ell)},\\
\Omega_\ell
&=
\left\{
(h,w):
0.15H_\ell \leq h < 0.85H_\ell,\;
0.15W_\ell \leq w < 0.85W_\ell
\right\},
\end{aligned}
\label{eq:center70}
\end{equation}
producing a fixed-length embedding $\mathbf{z}^{(\ell)}\in\mathbb{R}^{C_\ell}$. Global average pooling (GAP) uses the entire feature map instead of $\Omega_\ell$. The Center-70 embedding is used for abnormality scoring in Section~\ref{sec:score}. For CKA, the same central window is retained without average pooling so that the spatial organization of the feature map is preserved, as described in Section~\ref{sec:cka}.

\subsection{Normal-reference modeling and Mahalanobis abnormality scoring}
\label{sec:score}

For each split, only normal glomeruli from the fitting portion are used to standardize the embeddings and estimate the normal-reference distribution. Each embedding dimension is standardized to zero mean and unit variance using statistics estimated from the normal fitting embeddings. The same standardization parameters are then applied without re-estimation to all evaluation embeddings in that split. A multivariate Gaussian $\mathcal{N}(\boldsymbol{\mu},\boldsymbol{\Sigma})$ is fitted to the standardized normal embeddings.

Let $d$ denote the embedding dimension, where $d=C_\ell$ and $d=256$ for the selected $\mathrm{layer4}$ configuration, and let $n$ denote the number of normal glomeruli used for fitting. Across the five folds, $n$ ranges from $4{,}937$ to $5{,}996$. For the selected configuration, $n/d$ ranges from approximately $19.3$ to $23.4$. Although $n>d$, the embedding dimensions are strongly correlated, so the sample covariance matrix $\mathbf{S}$ can be poorly conditioned and its inverse unstable. We therefore use the Ledoit and Wolf shrinkage estimator~\cite{ledoit2022power}, as defined in Eq.~\eqref{eq:lw}:
\begin{equation}
\boldsymbol{\Sigma}
=
(1-\rho)\mathbf{S}
+
\rho\frac{\operatorname{tr}(\mathbf{S})}{d}\mathbf{I},
\label{eq:lw}
\end{equation}
where $\rho$ is the analytically determined shrinkage intensity and $\mathbf{I}$ is the $d\times d$ identity matrix. All abnormal categories are excluded when the standardization parameters, $\boldsymbol{\mu}$, and $\boldsymbol{\Sigma}$ are estimated.

An evaluation glomerulus is scored according to the distance between its standardized embedding and the normal reference distribution. The Mahalanobis abnormality score~\cite{ghorbani2019mahalanobis} is defined in Eq.~\eqref{eq:maha}:
\begin{equation}
d_M(\mathbf{z})
=
\sqrt{
(\mathbf{z}-\boldsymbol{\mu})^{\top}
\boldsymbol{\Sigma}^{-1}
(\mathbf{z}-\boldsymbol{\mu})
},
\label{eq:maha}
\end{equation}
where higher values indicate greater deviation from the normal feature distribution. We use $d_M$ rather than its squared value as the abnormality score. Because the two quantities are monotonically related, they produce identical rankings of the evaluated glomeruli.

\subsection{Layer-wise abnormality characterization using CKA}
\label{sec:cka}

The Mahalanobis distance indicates \emph{whether} a glomerulus is abnormal but not \emph{how} the abnormality changes the learned representation. To examine this distinction, we measure the similarity between network stages using linear CKA~\cite{lu2014multiple}. The analysis is performed within each image. For a single glomerulus, the feature map from each stage is cropped to the central window defined in Eq.~\eqref{eq:center70} and resampled to a common $32\times32$ spatial grid using average pooling. The resampled feature map is reshaped into an activation matrix $A^{(\ell)}\in\mathbb{R}^{n_s\times C_\ell}$, where $n_s=32\times32=1{,}024$. Each row represents one spatial location within the glomerulus, and each column represents one feature channel. Each column is mean-centered across the $n_s$ spatial locations, as required for linear CKA.

For a pair of stages $(\ell,\ell')$, with activation matrices $A^{(\ell)}\in\mathbb{R}^{n_s\times C_\ell}$ and $A^{(\ell')}\in\mathbb{R}^{n_s\times C_{\ell'}}$, linear CKA is calculated using Eq.~\eqref{eq:cka}:
\begin{equation}
\mathrm{CKA}\bigl(A^{(\ell)},A^{(\ell')}\bigr)
=
\frac{
\left\|
\bigl(A^{(\ell)}\bigr)^{\top}A^{(\ell')}
\right\|_F^2
}{
\left\|
\bigl(A^{(\ell)}\bigr)^{\top}A^{(\ell)}
\right\|_F
\left\|
\bigl(A^{(\ell')}\bigr)^{\top}A^{(\ell')}
\right\|_F
},
\label{eq:cka}
\end{equation}
where $\|\cdot\|_F$ denotes the Frobenius norm. This calculation yields one symmetric $11\times11$ matrix $M$ for each glomerulus. The rows and columns of $M$ both index the eleven Omni-Seg backbone stages. This procedure constitutes within-image spatial CKA because the aligned observations are the $1{,}024$ spatial locations within one glomerulus rather than different glomeruli.

The per-glomerulus matrices are aggregated by category. Specifically, $\bar{M}_c$ denotes the element-wise mean of $M$ across all glomeruli belonging to category $c$, and $\bar{M}_{\mathrm{normal}}$ denotes the element-wise mean across normal-reference glomeruli from the same institution. Each abnormality is then represented by the residual CKA signature defined in Eq.~\eqref{eq:resid}:
\begin{equation}
R_c(\ell,\ell')
=
\bar{M}_c(\ell,\ell')
-
\bar{M}_{\mathrm{normal}}(\ell,\ell').
\label{eq:resid}
\end{equation}
Positive values indicate that stages $\ell$ and $\ell'$ are more similar for abnormality category $c$ than for normal glomeruli, whereas negative values indicate that they are less similar. This analysis is descriptive and does not train an additional classifier; abnormality labels are used only to group glomeruli for averaging. The Mahalanobis score $d_M$ provides abnormality detection, whereas the residual CKA signature $R_c$ provides abnormality characterization.

\section{Data \& Experiments}

\subsection{Dataset Description}

Glomeruli were collected from periodic acid-Schiff-stained whole-slide images at two institutions. For both datasets, annotated glomerular regions with surrounding context were resampled to $1024\times1024$ pixels. No stain normalization or color augmentation was applied, and image intensities were scaled to $[0,1]$.

\noindent\textbf{Cornell dataset.} Slides were digitized at $0.2476~\mu\mathrm{m/pixel}$ using a $40\times$ objective. The source category \emph{Viable glomeruli} was confirmed to represent normal glomeruli and was used as the normal reference. The abnormal categories were global glomerulosclerosis, ischemic, segmental glomerulosclerosis, and atubular glomeruli.

\noindent\textbf{Vanderbilt dataset.} Slides were digitized at $0.25~\mu\mathrm{m/pixel}$ using a $40\times$ objective. The source category \emph{Normal glomeruli} was used as the normal reference. The abnormal categories were obsolescent, solidified, and disappearing glomeruli.

Only glomerular patches were included in the analysis. The two normal categories were pooled to form the negative class for abnormality detection, while all remaining categories were treated as positive. Dataset composition is summarized in Table~\ref{tab:data}. Abnormal labels were not used to fit the normal-reference model; they were used for configuration selection, evaluation, and grouping glomeruli for CKA analysis. The analyzed glomeruli come from 153 whole-slide images: 17 from Cornell and 136 from Vanderbilt. Each slide contains multiple glomeruli and can contribute glomeruli to more than one category, so the slide counts are not additive across the categories in Table~\ref{tab:data}. The two cohorts also differ in how the data were collected. Every Cornell slide contains viable glomeruli, and nearly all slides contribute to each abnormal category. In contrast, for the Vanderbilt slides, normal glomeruli were labeled from only a small portion of slides, while the abnormal categories come from a much larger set of slides.

\begin{table}[t]
\centering
\small
\caption{Dataset composition. The two reference categories were pooled to form the negative class and were the only categories used to fit the normal-reference model. The remaining categories were used for configuration selection, evaluation, and CKA analysis.}
\label{tab:data}
\begin{tabular}{llr}
\toprule
Institution & Category & \# glomeruli \\
\midrule
Cornell & Normal glomeruli (source label: Viable) & 3{,}223 \\
Cornell & Global glomerulosclerosis & 2{,}500 \\
Cornell & Ischemic & 502 \\
Cornell & Segmental glomerulosclerosis & 271 \\
Cornell & Atubular & 230 \\
\midrule
Vanderbilt & Normal glomeruli & 3{,}609 \\
Vanderbilt & Obsolescent (Global glomerulosclerosis) & 4{,}289 \\
Vanderbilt & Solidified (Global glomerulosclerosis) & 725 \\
Vanderbilt & Disappearing (Global glomerulosclerosis) & 434 \\
\bottomrule
\end{tabular}
\end{table}

\subsection{Experimental Details}

\noindent\textbf{Normal-reference fitting and data splitting.} All splits were performed at the whole-slide image level so that glomeruli from the same slide were not shared between model fitting and evaluation. Slides were sorted, randomly shuffled, and then evenly distributed within each institution, so the split was matched by institution. Each patient in the normal cohort contributed one slide, making the slide-level and patient-level splits equivalent. The 40 normal slides from both institutions were pooled and divided into five fixed folds containing 9, 9, 8, 7, and 7 slides. In each experiment, four folds were used to fit the normal-reference model, while the remaining fold provided the held-out normal evaluation set. Every abnormal glomerulus was evaluated against each of the five fold-specific reference models. Thus, the folds shared the same abnormal cases and differed only in their normal fitting and evaluation sets.

\noindent\textbf{Configuration selection and final detection.} The eleven Omni-Seg stages were evaluated individually using GAP, and spatial pooling configurations were compared using $\mathrm{x1\_resb}$. Based on these comparisons, Center-70 was combined with the selected $\mathrm{layer4}$ representation, producing the final $256$-dimensional embedding used for detection and baseline comparisons. Layer-pooling interactions were not exhaustively evaluated. Because abnormal labels were used to compare configurations and the same abnormal cases were included in the final evaluation, the reported results are post-selection estimates rather than results from an independent test cohort. The pipeline is described as training-free because the Omni-Seg backbone receives no gradient updates and only normal glomeruli are used to estimate the standardization statistics and Gaussian normal-reference model.

\noindent\textbf{CKA analysis.} Residual CKA was calculated separately for each institution using its corresponding normal reference, as defined in Eq.~\eqref{eq:resid}. All eleven Omni-Seg stages were evaluated using the central-window processing described in Section~\ref{sec:cka}. 

\noindent\textbf{Hardware and inference.} All experiments were run on a single NVIDIA RTX 6000 Ada Generation GPU (48 GB) with an Intel Xeon w9-3475X CPU, using PyTorch 2.12 with CUDA 13.0 and scikit-learn 1.7. The normal-reference model does not require gradient computation. Standardization and the Ledoit–Wolf covariance estimate are both closed-form operations and take only a few seconds per fold on the CPU. As a result, the main computational cost comes from the single feature-extraction pass.

\subsection{Comparison Methods}

NoRDeC was compared with PaDiM~\cite{defard2021padim} and PatchCore~\cite{roth2022towards} using the same frozen Omni-Seg backbone, five data splits, approximately matched central regions, and evaluation metrics. NoRDeC uses the $256$-dimensional Center-70 pooled embedding from $\mathrm{layer4}$. The baselines use position-wise features from $\mathrm{layer2}$, $\mathrm{layer3}$, and $\mathrm{layer4}$, reduced to 100 channels on a $16\times16$ spatial grid. Scoring is restricted to the central $12\times12$ positions, which provide the closest discrete approximation to Center-70 on this grid. The comparison therefore evaluates complete scoring strategies under a shared backbone and approximately matched central regions rather than identical feature tensors.

\subsection{Evaluation Metrics}

Detection performance was evaluated using the area under the receiver operating characteristic curve (AUROC) and the area under the precision-recall curve (AUPRC), with abnormal glomeruli treated as the positive class. Both metrics are reported as the mean and standard deviation across the five folds. Differences between NoRDeC and each baseline were assessed using paired \(t\)-tests of the five fold-level AUROC estimates. The resulting standard deviations and statistical comparisons are reported in Table~\ref{tab:overall_comparison}.

Because the abnormality categories differ in prevalence, AUPRC must be interpreted relative to its chance level, which equals the positive-class prevalence. Table~\ref{tab:percat} reports the mean fold-specific chance AUPRC for each abnormality category, whereas Table~\ref{tab:overall_comparison} reports the pooled chance AUPRC of \(0.869\). Raw AUPRC values are therefore not directly comparable across categories. NoRDeC produces the continuous Mahalanobis abnormality score \(d_M\), and AUROC and AUPRC evaluate the ranking performance of this score.

\begin{figure}[t]
    \centering
    \includegraphics[width=0.8\linewidth]{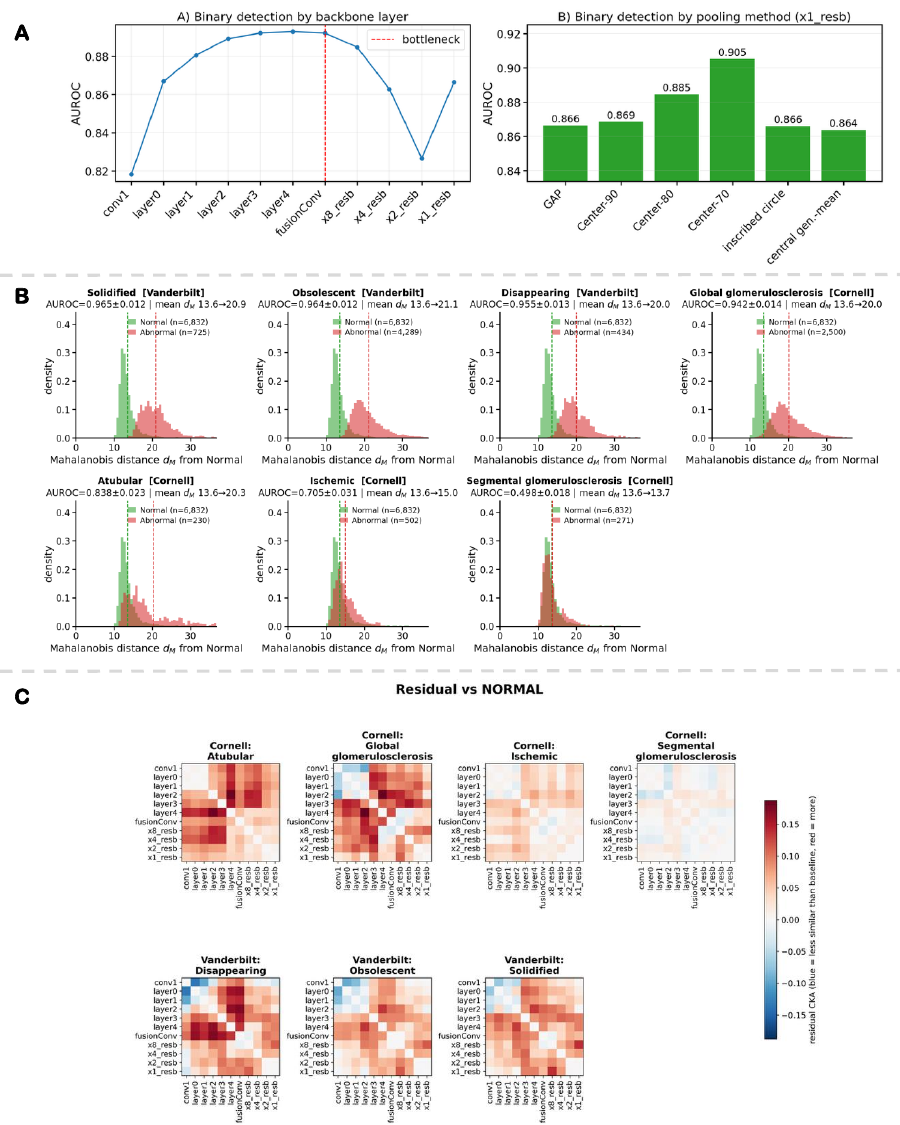}
    \caption{Summary of abnormality detection and representation characterization. (A) Detection performance across Omni-Seg backbone stages and spatial aggregation strategies. (B) Mahalanobis-distance distributions for normal and abnormal glomeruli within each abnormality-specific evaluation. Dashed lines indicate the corresponding mean distances. (C) Residual CKA matrices showing changes in inter-stage similarity relative to the institution-specific normal reference. Red indicates greater inter-stage similarity, whereas blue indicates lower inter-stage similarity relative to normal glomeruli. All residual CKA matrices use the same color scale.}
    \label{fig:results_overview}
\end{figure}

\section{Results}

The principal results are summarized in Fig.~\ref{fig:results_overview}, including abnormality-score distributions, the effects of backbone layer and spatial aggregation, and residual CKA signatures.

\subsection{Influence of backbone layer and spatial aggregation on detection performance}
\label{sec:res_layer}

Detection performance depended on both the backbone stage and spatial aggregation strategy, as shown in Fig.~\ref{fig:results_overview}A and Table~\ref{tab:configuration_results}. Using GAP, AUROC increased from $0.818$ at $\mathrm{conv1}$ to $0.893$ at $\mathrm{layer4}$ and subsequently declined across the decoder. These results indicate that the abnormality signal is most strongly represented in the intermediate-to-deep encoder stages.

Restricting spatial aggregation to the central region also improved detection. At $\mathrm{x1\_resb}$, AUROC increased from $0.866$ with GAP to $0.905$ with Center-70. The other evaluated aggregation strategies did not exceed Center-70. Combining the selected $\mathrm{layer4}$ representation with Center-70 produced a pooled AUROC of $0.926$. This configuration was therefore used for the subsequent detection analyses.

\begin{table}[t]
\centering
\small
\caption{Detection performance for the principal layer and aggregation configurations. Values are reported as mean $\pm$ standard deviation across five folds. Bold indicates the best mean AUROC within each analysis.}
\label{tab:configuration_results}

\begin{tabular}{llr@{\,${}\pm{}$\,}l}
\toprule
Analysis
& Configuration
& \multicolumn{2}{c}{AUROC} \\
\midrule

Backbone stage
& $\mathrm{conv1}$ with GAP
& 0.818 & 0.045 \\

Backbone stage
& $\mathrm{layer4}$ with GAP
& \textbf{0.893} & 0.013 \\

Backbone stage
& $\mathrm{fusionConv}$ with GAP
& 0.892 & 0.013 \\

Backbone stage
& $\mathrm{x2\_resb}$ with GAP
& 0.827 & 0.026 \\

\midrule

Spatial aggregation
& $\mathrm{x1\_resb}$ with GAP
& 0.866 & 0.024 \\

Spatial aggregation
& $\mathrm{x1\_resb}$ with Center-90
& 0.869 & 0.024 \\

Spatial aggregation
& $\mathrm{x1\_resb}$ with Center-80
& 0.885 & 0.021 \\

Spatial aggregation
& $\mathrm{x1\_resb}$ with Center-70
& \textbf{0.905} & 0.018 \\

Spatial aggregation
& $\mathrm{x1\_resb}$ with inscribed-circle pooling
& 0.866 & 0.023 \\

Spatial aggregation
& $\mathrm{x1\_resb}$ with central generalized-mean pooling
& 0.864 & 0.021 \\

\bottomrule
\end{tabular}
\end{table}

\subsection{Overall and category-specific abnormality detection performance}
\label{sec:res_cat}

The Mahalanobis-distance distributions in Fig.~\ref{fig:results_overview}B show that most abnormality categories were shifted toward larger distances relative to normal glomeruli. Table~\ref{tab:percat} reports the corresponding category-specific AUROC and AUPRC values for NoRDeC, PaDiM, and PatchCore, together with the chance AUPRC for each category.

NoRDeC achieved the highest pooled AUROC of \(0.926\), compared with \(0.830\) for PaDiM~\cite{defard2021padim} and \(0.752\) for PatchCore~\cite{roth2022towards}. NoRDeC also achieved the highest pooled AUPRC of \(0.982\), compared with \(0.959\) for PaDiM and \(0.937\) for PatchCore. The corresponding standard deviations and fold-level statistical comparisons are reported in Table~\ref{tab:overall_comparison}. Because the pooled chance AUPRC was \(0.869\), the pooled AUPRC values should be interpreted relative to the high prevalence of abnormal glomeruli in the evaluation set.

\begin{table}[t]
\centering
\small
\setlength{\tabcolsep}{2.5pt}
\caption{Per-category abnormality detection performance. Model performance is reported as mean $\pm$ standard deviation across five folds. Chance AUPRC is the mean fold-specific positive prevalence. Bold indicates the best mean AUROC and AUPRC within each row.}
\label{tab:percat}

\begin{adjustbox}{max width=\linewidth}
\begin{tabular}{
l
c
r@{\,${}\pm{}$\,}l
r@{\,${}\pm{}$\,}l
r@{\,${}\pm{}$\,}l
r@{\,${}\pm{}$\,}l
r@{\,${}\pm{}$\,}l
r@{\,${}\pm{}$\,}l
}
\toprule
&
\multirow{2}{*}{\makecell{Chance\\AUPRC}}
& \multicolumn{4}{c}{NoRDeC (Ours)}
& \multicolumn{4}{c}{PaDiM~\cite{defard2021padim}}
& \multicolumn{4}{c}{PatchCore~\cite{roth2022towards}} \\
\cmidrule(lr){3-6}
\cmidrule(lr){7-10}
\cmidrule(lr){11-14}

Category
&
& \multicolumn{2}{c}{AUROC}
& \multicolumn{2}{c}{AUPRC}
& \multicolumn{2}{c}{AUROC}
& \multicolumn{2}{c}{AUPRC}
& \multicolumn{2}{c}{AUROC}
& \multicolumn{2}{c}{AUPRC} \\
\midrule

Solidified
& 0.358
& \textbf{0.965} & 0.012
& \textbf{0.884} & 0.077
& 0.807 & 0.034
& 0.608 & 0.092
& 0.794 & 0.028
& 0.603 & 0.077 \\

Obsolescent
& 0.761
& \textbf{0.964} & 0.012
& \textbf{0.977} & 0.018
& 0.842 & 0.030
& 0.918 & 0.029
& 0.804 & 0.025
& 0.902 & 0.029 \\

Disappearing
& 0.251
& \textbf{0.955} & 0.013
& \textbf{0.795} & 0.106
& 0.843 & 0.029
& 0.581 & 0.096
& 0.769 & 0.023
& 0.463 & 0.075 \\

Global glomerulosclerosis
& 0.652
& \textbf{0.942} & 0.014
& \textbf{0.947} & 0.035
& 0.882 & 0.024
& 0.910 & 0.032
& 0.709 & 0.034
& 0.769 & 0.052 \\

Atubular
& 0.152
& \textbf{0.838} & 0.023
& \textbf{0.570} & 0.091
& 0.810 & 0.033
& 0.420 & 0.088
& 0.718 & 0.032
& 0.274 & 0.054 \\

Ischemic
& 0.279
& \textbf{0.705} & 0.031
& \textbf{0.448} & 0.078
& 0.643 & 0.042
& 0.418 & 0.077
& 0.598 & 0.044
& 0.328 & 0.066 \\

Segmental glomerulosclerosis
& 0.174
& 0.498 & 0.018
& 0.196 & 0.045
& \textbf{0.562} & 0.039
& \textbf{0.253} & 0.057
& 0.497 & 0.038
& 0.169 & 0.035 \\

\bottomrule
\end{tabular}
\end{adjustbox}
\end{table}

\begin{table}[t]
\centering
\small
\caption{Overall abnormality detection performance across five folds. Metric values are reported as mean $\pm$ standard deviation, and the pooled chance AUPRC is $0.869$. Statistical comparisons are paired $t$-tests of fold-level AUROC between NoRDeC and each baseline. Bold indicates the best mean value for each metric.}
\label{tab:overall_comparison}

\begin{tabular}{
l
r@{\,${}\pm{}$\,}l
r@{\,${}\pm{}$\,}l
l
}
\toprule
Method
& \multicolumn{2}{c}{AUROC}
& \multicolumn{2}{c}{AUPRC}
& Statistical comparison \\
\midrule

NoRDeC (Ours)
& \textbf{0.926} & 0.013
& \textbf{0.982} & 0.011
& Reference \\

PaDiM~\cite{defard2021padim}
& 0.830 & 0.029
& 0.959 & 0.015
& $t(4)=10.1,\ p<0.001$ \\

PatchCore~\cite{roth2022towards}
& 0.752 & 0.029
& 0.937 & 0.019
& $t(4)=17.7,\ p<0.001$ \\

\bottomrule
\end{tabular}
\end{table}

Performance differed substantially among abnormality categories. The solidified, obsolescent, disappearing, and global glomerulosclerosis categories achieved AUROCs from $0.942$ to $0.965$. Atubular glomeruli achieved an AUROC of $0.838$, whereas ischemic glomeruli achieved a lower AUROC of $0.705$. These differences may reflect the greater morphological overlap between subtle abnormalities and normal glomeruli.

Table~\ref{tab:distance} shows the corresponding Mahalanobis distances. The solidified, obsolescent, disappearing, and global glomerulosclerosis categories were all similarly distant from the normal reference, even though they represent different pathological processes. This suggests that the Mahalanobis score captures how far a glomerulus has deviated from normal morphology rather than distinguishing between specific types of abnormalities. In contrast, ischemic and segmental glomerulosclerosis glomeruli remained relatively close to the normal mean. Detection also depended on how much variation there was within each category, not just on how far the category was displaced from normal. For example, atubular glomeruli were just as far from the normal reference as the global glomerulosclerosis category, but they showed much greater variability. As a result, a larger proportion of atubular glomeruli still fell within the normal range.

Segmental glomerulosclerosis was the most challenging category, with NoRDeC producing an AUROC of $0.498$. PaDiM produced a numerically higher AUROC of $0.562$, while PatchCore achieved $0.497$. This was the only category for which NoRDeC did not produce the highest AUROC or AUPRC. The result suggests that position-wise modeling may be advantageous for abnormalities confined to a limited portion of the glomerulus.

\subsection{Layer-wise representation characterization using CKA}
\label{sec:res_cka}

The Mahalanobis score indicates the extent to which a glomerulus deviates from the normal reference but does not describe how the relationships between backbone stages differ. Residual CKA provides this complementary representation characterization, as shown in Fig.~\ref{fig:results_overview}C.

Global glomerulosclerosis and atubular glomeruli produced their largest residuals between $\mathrm{layer2}$ and $\mathrm{layer4}$, with values of $+0.187$ and $+0.182$, respectively. Disappearing glomeruli produced their largest residual between $\mathrm{layer2}$ and $\mathrm{fusionConv}$ ($+0.173$), whereas solidified glomeruli produced their largest residual between $\mathrm{x8\_resb}$ and $\mathrm{x1\_resb}$ ($+0.139$). Segmental glomerulosclerosis produced comparatively small residuals, with a maximum absolute residual of $0.042$, consistent with its limited separation from normal glomeruli.

Obsolescent and solidified glomeruli achieved similar AUROCs of $0.964$ and $0.965$, respectively, but produced different residual CKA patterns. Obsolescent glomeruli primarily altered relationships involving the intermediate and deep encoder stages, whereas solidified glomeruli produced stronger changes within the decoder. Thus, abnormalities with similar detection performance can produce different layer-wise representation signatures. Because abnormality categories differ between the two institutions, these signatures should be interpreted relative to their corresponding institution-specific normal references rather than as institution-independent biological effects.

\begin{table}[t]
\centering
\small
\caption{Mahalanobis distance from the normal reference by category using \(\mathrm{layer4}\) with Center-70 pooling. The reported values are the mean \(\pm\) standard deviation of the fold-specific \(d_M\) scores. The column \(n\) gives the number of unique glomeruli. Each abnormal glomerulus was scored against all five fold-specific normal-reference models and therefore contributed five scores to its category summary. Each normal glomerulus contributed the score obtained when its slide was included in the held-out fold.}
\label{tab:distance}

\begin{tabular}{llrc}
\toprule
\textbf{Category}
& \textbf{Institution}
& \(n\)
& \textbf{Mean \(d_M\)} \\
\midrule

Normal (held-out reference)
& Both
& 6{,}832
& \(13.55 \pm 2.84\) \\

\midrule

Solidified
& Vanderbilt
& 725
& \(20.91 \pm 4.05\) \\

Obsolescent
& Vanderbilt
& 4{,}289
& \(21.13 \pm 4.66\) \\

Disappearing
& Vanderbilt
& 434
& \(20.00 \pm 3.73\) \\

Global glomerulosclerosis
& Cornell
& 2{,}500
& \(20.05 \pm 4.33\) \\

Atubular
& Cornell
& 230
& \(20.29 \pm 9.24\) \\

Ischemic
& Cornell
& 502
& \(14.99 \pm 2.65\) \\

Segmental glomerulosclerosis
& Cornell
& 271
& \(13.68 \pm 2.71\) \\

\bottomrule
\end{tabular}
\end{table}

\subsection{Limitations}
\label{sec:limits}

The backbone stage and spatial aggregation strategy were selected using the same folds used for evaluation because no separate validation cohort was available, which may overestimate performance on independent data. Normal glomeruli from both institutions were pooled without stain normalization, while each abnormality category was obtained from only one institution; category and institution are therefore confounded. For the Vanderbilt and Cornell datasets, the normal reference was built from different numbers of slides. The normal glomeruli for Vanderbilt were labeled on only a small number of slides, but then it was used to evaluate glomeruli from many more slides. For Cornell, normal glomeruli come from nearly all slides. Because of this, the Vanderbilt reference may not capture the full range of normal appearances in the cohort, which could make otherwise normal glomeruli from unseen slides appear more unusual. PaDiM and PatchCore were evaluated using one fixed configuration each, so the comparisons apply only to the evaluated settings. Finally, the present study analyzes only glomeruli. Expanding the normal annotations to more slides, studying how the reference changes with greater slide coverage, and extending normal-reference detection and representation analysis to other renal structures, including tubules, is an important direction for future work.

\section{new or breakthrough work to be presented}

NoRDeC combines normal-reference abnormality scoring with layer-wise representation characterization using a frozen segmentation backbone pretrained without abnormality supervision. The backbone receives no gradient updates, and only normal glomeruli are used to estimate the standardization statistics and Gaussian normal-reference distribution. Abnormality labels are used for configuration selection, evaluation, and grouping glomeruli for residual CKA analysis. Under the evaluated configurations, NoRDeC achieved the highest pooled AUROC among the three methods while using the same frozen backbone and data splits and approximately matched central regions. Beyond detection, residual CKA identifies distinct layer-wise representation signatures for different abnormality categories relative to their corresponding institution-specific normal references.

\section{Conclusion}

We presented NoRDeC, a framework for detecting and characterizing glomerular abnormalities using frozen feature representations learned without abnormality supervision. NoRDeC achieved a pooled AUROC of $0.926$, exceeding the evaluated PaDiM and PatchCore configurations. Residual CKA further showed that abnormality categories with similar detection performance can produce different layer-wise representation signatures relative to their institution-specific normal references. Detection of segmental glomerulosclerosis remained near chance, possibly because spatial pooling reduces sensitivity to changes confined to a limited portion of the glomerulus. These findings support normal-reference feature analysis as a framework for glomerular abnormality detection and representation characterization while motivating independent multi-institutional validation and more localized scoring strategies.

\section{ACKNOWLEDGMENTS} 
This research was supported by the WCM Radiology AIMI Fellowship, WCM CTSC 2027 Pilot Award, NIH 1R01HL174863-01A1 (Sabuncu) and 1U54DK144866-01 (Sabuncu).

\bibliography{main} 
\bibliographystyle{spiebib} 

\end{document}